\documentclass{article}

\PassOptionsToPackage{numbers, compress}{natbib}

\usepackage[final]{nips_2017}
\usepackage{amsmath}
\usepackage[utf8]{inputenc} % allow utf-8 input
\usepackage[T1]{fontenc}    % use 8-bit T1 fonts
\usepackage{hyperref}       % hyperlinks
\usepackage{url}            % simple URL typesetting
\usepackage{booktabs}       % professional-quality tables
\usepackage{amsfonts}       % blackboard math symbols
\usepackage{nicefrac}       % compact symbols for 1/2, etc.
\usepackage{microtype}      % microtypography
\usepackage{graphicx}
\usepackage{caption}
\usepackage{wrapfig}

\title{Personalized Driver Stress Detection with Multi-task Neural Networks using Physiological Signals}

\author{
  Aaqib Saeed\\ Eindhoven University of Technology \\ Eindhoven, The Netherlands \\ \texttt{a.saeed@tue.nl}
  \And Stojan Trajanovski \\ Philips Research \\ Eindhoven, The Netherlands \\ \texttt{stojan.trajanovski@philips.com}
}

\begin{document}

\maketitle

\begin{abstract}
Stress can be seen as a physiological response to everyday emotional, mental and physical challenges. A long-term exposure to stressful situations can have negative health consequences, such as increased risk of cardiovascular diseases and immune system disorder. Therefore, a timely stress detection can lead to systems for better management and prevention in future circumstances. In this paper, we suggest a multi-task learning based neural network approach (with hard parameter sharing of mutual representation and task-specific layers) for personalized stress recognition using skin conductance and heart rate from wearable devices. The proposed method is tested on multi-modal physiological responses collected during real-world and simulator driving tasks. 
\end{abstract}

\section{Introduction}

Stress is described as a physiological response to emotional, mental and physical challenges, which people face in everyday life~\cite{schneiderman2005stress}. Numerous types of stressors are part of today's modern life, such as exams or annual job evaluations. Even though human body adjusts with day-to-day stressors, the long-term exposure to extreme stress can be destructive for mental as well as physical health~\cite{pickering2001mental}. It also increases the risk of cardiovascular diseases and (psycho) somatic complaints~\cite{holmes2006mental,shi2007galvanic}. Due to the health issues associated with stress, its measurement and management become important. A timely detection of stress can help users to take corrective and preventive measures in an informed way. 

The physiological stress affects two branches of an autonomic nervous system: sympathetic nervous system and parasympathetic nervous system. The immediate effect of their stimulations is a measurable change in physiological parameters, such as an increased heart rate (HR) and skin conductance level~\cite{dawson2007electrodermal}. Stress research has a wide area of applications, from increasing resilience of military personnel to improving athletes' performance. Many techniques have been proposed in the past to detect stress in pilots~\cite{sem1961electroencephalographic}, car drivers~\cite{healey2005detecting,hennessy1999traffic}, computer users~\cite{zhai2006stress}, and in surgeons~\cite{sexton2000error}. In addition to speech and facial expressions, most of the approaches use numerous physiological signals~\cite{sharma2012objective}, such as respiration rate, electrocardiography (ECG), blood pressure, and electromyography (EMG). The collection of these data sources in naturalistic conditions is very difficult and not consumer friendly for developing practical applications. On the contrary, skin conductance and heart rate can be reliably acquired in a non-invasive way from wearable sensors placed on the wrist.

In this paper, we focus on stress detection (binary classification) during real-world and simulated driving tasks using skin conductance and heart rate data. The physiological signals tend to vary in people which are influenced by age, gender, diet or sleep~\cite{picard2001toward}. Due to this fact, stress responses can differ from person to person. The global (or one-fits-all) models that are usually used~\cite{healey2005detecting} for stress recognition, often do not generalize well to unseen test subjects and hence require extensive fine-tuning of the model. Therefore, to take account of the interpersonal differences, we adopt a multi-task learning (MTL) approach with subject-as-tasks. Specifically, the MTL model has hard parameter sharing of mutual representation along with a specific layer for each subject (or task) in order to personalize the model.  

The main contribution of this paper is to use multi-modal physiological data of real-world and simulator driving to develop a multi-task neural network for personalized stress detection. 

\section{Dataset and Feature Extraction}

\subsection{MIT Driver Stress}

The MIT Driver Stress dataset~\cite{healey2002driver} consists of physiological signals recorded during a real-life experiment with subjects in following conditions: 1) resting, 2) driving in a city, and 3) on a highway. The dataset consists of $17$ drives, where each driving session lasts for $1$-$1.5$ hours. The recorded signals are EMG, ECG, galvanic skin response (GSR) from hand and foot, heart rate derived from ECG and respiration rate. The GSR and respiration rate is sampled at $31$ Hz, ECG recorded at $496$ Hz and EMG at $15.5$ Hz. The signals provided in the dataset are down-sampled to $15.5$ Hz. There is another signal available in the dataset called ‘marker’. It indicates a change of activity (a button press) i.e. the start or end of a rest period, city or highway driving. 

The marker signal is used to derive ground truth annotation for binary stress levels. Peaks are detected in the signal to capture the button push event; indicating a new trial of the experiment is commencing. The data points before and after the first and last marker (peaks) are removed as they correspond to the time when subjects were equipped with sensors. Likewise, $4$ minutes of data after resting and before the beginning of post driving baseline are removed. These steps are taken to avoid feeding signals with ambiguous labels, as it is hard to determine if users are stressful or recuperated. The artifacts are removed from HR and GSR signals following~\cite{ollander2015wearable} as values fluctuated to unreasonably high and low. The EMG signal is discarded because the sensor placed on the shoulder and it might have recorded muscle movement instead of psychological stress response~\cite{ollander2015wearable}. Likewise, ECG, GSR from foot and respiration rate are also not used as collecting this data in real-world settings is very problematic. Finally, the following $10$ drives’ dataset having GSR from hand, HR and marker signals available are used for model training and evaluation: $04$, $05$, $06$, $07$, $08$, $09$, $10$, $11$, $12$ and $16$. 

\subsection{Simulator Driving}
We collected heart rate and skin conductance (SC) data from $19$ professional truck drivers using wrist-worn devices. The SC signal was recorded at a frequency of $10$ Hz and HR was derived from Photoplethysmogram sensor data with a frequency of $1$ Hz; it is upsampled to match the frequency of SC. The experiment was realized with a driving simulation software and participants received standardized instructions from an audiotape. The high stress was induced by means of secondary arithmetic subtraction task. It is a component of widely used Trier Social Stress Test~\cite{birkett2011trier}, where a user has to perform serial subtraction verbally in a loud manner and have to start over from the last correct answer; if a mistake is made.

The study consisted of three major steps 1) baseline driving, 2) moderate stress activity, and 3) high-stress task~\citep{saeed2017ddeep}. The experimental trial was initiated with a normal driving for $15$ minutes. Afterwards, each subject was asked twice to count $1$-$60$ as a moderate stress activity with a very small interval between two activities. After a one-minute period of normal driving and to induce high stress, the subject was asked to count backward from a random number in steps of $7$ in approximately $30$ seconds. Subsequently, the subject was asked to count backward again from another random number. This process was repeated for approximately $5$ minutes. The length of the stress simulation task was $25$ minutes, including baseline. Since we were interested in recognition of baseline and high stress, data points of moderate stress activity and bad quality signals of two subjects were dropped.

\subsection{Features}
For model input, we used a sliding window approach to extract physiological features from each participant’s data. A similar window length of $30$ seconds with a fixed step size of $15$ seconds (or $50$\% overlap) is used for both (real-world and simulator) datasets. It is important to note that, features were computed from pre-processed signals, and were subsequently standardized with mean normalization by baseline to compensate for individuals having different resting heart rates. 

\subsubsection*{Heart Rate}
The heart rate measures the number of heartbeats per unit of time. It describes the heart activity when the autonomic nervous system attempts to tackle with the human body's demands depending on the received stimuli~\cite{healey2005detecting}. We obtained the following seven features from heart rate: mean, standard deviation, min, max, range, root mean square of successive differences, and standard deviation of successive differences.

\subsubsection*{Skin Conductance}
The skin conductance (also known as galvanic skin response) describes the autonomic variations in electrical properties of the skin or equivalently, number of active sweat glands. It is widely used as a sensitive index of emotional processing, sympathetic activity and is a relevant indicator of the stress level of a person~\cite{labbe2007coping,ferreira2008license}. From this signal the following nine features are extracted: mean, standard deviation, min, max, range, number of peaks, amplitude, skewness, and kurtosis. 

\section{Stress Classification}

\subsection{Problem Formulation}
The stress detection task can be formulated as a supervised sequence classification problem. In this task, the objective is to assign a single label to an input sequence. It can be conceived as follows, let $\{( \pmb {x_i},y_i)\}^m_{i=1}$ be a dataset with $m$ sequences of fixed length. The high-level features ($16$ features in our case) are computed from each raw input sequence $\pmb{x_i}$ and corresponding label $y_i$ (binary label in this case) is generally assigned to be the mode of context window labels.  

\subsection{Model Architecture}

\begin{wrapfigure}{r}{0.47\textwidth}
\begin{center}
  \includegraphics[width = \linewidth]{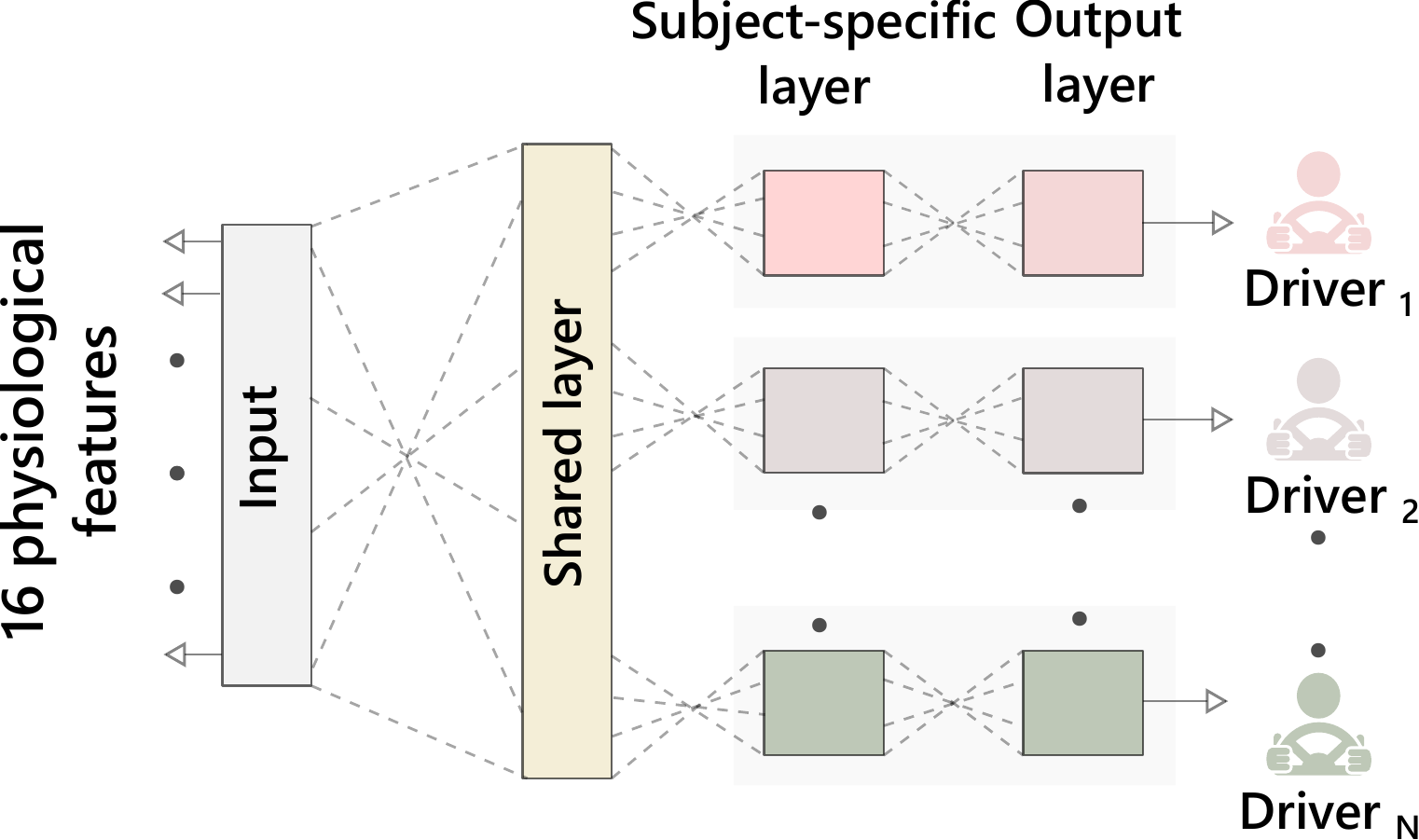}
  \captionsetup{font=scriptsize,labelfont=scriptsize}
  \caption{Multi-task neural network architecture with one shared layer (hard-parameter sharing) and a subject (driver) specific dense layer with sigmoid classifiers in the last layer. The model input is a vector of $16$ physiological features, extracted from heart rate and skin conductance (see section 2.3) using windows of length $30$ seconds. The same output from shared layer is fed into subject-specific layers for personalization.}
  \label{fig:mtl}
\end{center}
\end{wrapfigure}

The neural network learns complex non-linear transformations of the input data through several hidden layers, having a different number of neurons connected together. In a single-task neural network (ST-NN), there is only one task to solve by minimizing a single loss function with backpropagation. Conversely, multi-task learning involves finding a unified model for solving more than one task with a shared representation of the tasks. Consequently, multi-task neural network model (MT-NN) consists of common layers mutual across tasks as well as task-specific layers. Moreover, in the last layer, there is a separate $sigmoid$ unit and a loss function for each task. The optimization of loss functions is done simultaneously at random or in other words, by alternating between different tasks. 

The multi-task learning is generally done through hard or soft parameter sharing~\cite{ruder2017overview}. By following Jaques et al.~\cite{jaquesmulti}, we employed hard parameter sharing, where, final layers are subject-specific as shown in Figure~\ref{fig:mtl}. We used a shared fully-connected layer with $200$ neurons and $elu$ $( \alpha \ (exp(x) -1)$ if $x < 0$ else $x)$ activation. The subject-specific layers have $50$ neurons and $elu$ to reflect non-linearity. Likewise, we applied l2-regularization on task-specific layers and validation-based early stopping, to avoid over-fitting. The binary cross-entropy is optimized as an objective function using a variant of stochastic gradient descent `Adam'~\cite{kingma2014adam}. This model architecture will be able to take interpersonal variations in physiological signals into account through person-specific layers while having a mutual global representation. 

\section{Results}
Our experiments were conducted using a) MIT Driver Stress dataset~\cite{healey2002driver} and b) Simulator Driving data. We first tested two standard classifiers as a baseline: logistic regression (LR) and support vector machine with linear (L) and radial basis function (RBF) kernels. In addition to that, we also trained two layers (subject independent) neural network model for performance comparison. The data of each subject is divided randomly into train and test sets ($80$/$20$). The $5$-fold cross validation is performed on the training set for hyper-parameter optimization and evaluation metrics are averaged across participants on the test set. The stress recognition performance of these models is summarized in Table~\ref{tab:on_road} and~\ref{tab:simulator} for real-world and simulator driving, respectively. 

\begin{table}[hbp]
\centering
\setlength\tabcolsep{4pt}
\begin{minipage}{0.48\textwidth}
\centering
\captionsetup{font=small,labelfont=small}
\caption{Average test set ($20$\%) results of subjects (or drives) in MIT Driver Stress dataset}
\begin{small}
\begin{tabular}{lll}
\hline
\multicolumn{1}{c}{\textbf{Model}} & \multicolumn{1}{c}{\textbf{F-Score}} & \multicolumn{1}{c}{\textbf{Kappa}} \\ \hline
LR & 0.894 $\pm$ 0.078 & 0.672 $\pm$ 0.191 \\
SVM (L) & 0.903 $\pm$ 0.076 & 0.706 $\pm$ 0.175 \\
SVM (RBF) & 0.950 $\pm$ 0.027 & 0.828 $\pm$ 0.100 \\
ST-NN & 0.954 $\pm$ 0.027 & 0.844 $\pm$ 0.095 \\
MT-NN & \textbf{0.965} $\pm$ \textbf{0.023} & \textbf{0.879} $\pm$ \textbf{0.080} \\ \hline
\end{tabular}
\end{small}
\label{tab:on_road} 
\end{minipage}
\hfill
\begin{minipage}{0.48\textwidth}
\centering
\captionsetup{font=small,labelfont=small}
\caption{Average test set ($20$\%) results of participants of Simulator Driving dataset} 
\small{
\begin{tabular}{lll}
\hline
\multicolumn{1}{c}{\textbf{Model}} & \multicolumn{1}{c}{\textbf{F-Score}} & \multicolumn{1}{c}{\textbf{Kappa}} \\ \hline
LR & 0.720 $\pm$  0.342 & 0.663 $\pm$  0.371 \\
SVM (L) & 0.726 $\pm$  0.326 & 0.671 $\pm$  0.349 \\
SVM (RBF) & 0.774 $\pm$ 0.300 & 0.710 $\pm$  0.371 \\
ST-NN & 0.801 $\pm$  0.243 & 0.736 $\pm$  0.307 \\
MT-NN & \textbf{0.922} $\pm$ \textbf{0.137} & \textbf{0.891} $\pm$  \textbf{0.184} \\ \hline
\end{tabular}
}
\label{tab:simulator} 
\end{minipage}
\end{table}

In case of the on-road dataset, it can be seen that mostly all models performed well but there is a considerable variance in the results of standard classifiers. The ST-NN model reduced the spread achieving average f-score and kappa of $0.95$ and $0.84$, respectively. Likewise, MT-NN model improved the results even further by minimizing the std. deviation and reach an f-score value of $0.96$ and $0.88$ of kappa. It can be seen as an overall improvement across drives due to subject-specific layers. However, caution is advised in the interpretation of MIT Driver Stress dataset's result as no actual ground truth annotations or subjective self-reports are publicly available. The labels were acquired by means of a `marker' signal, representing the start of next study trial (i.e. from resting to driving in a city) and assuming that driving, in general, is a stressful task. For simulator driving, the commonly used classifiers and ST-NN do not generalize well as can be noticed from huge standard deviation values of evaluation metrics. The MT-NN notably improved the recognition rate across subjects and resulted in a better model by achieving f-score and kappa of $0.92$ and $0.89$, respectively. We think the reason for the large variation in results across participants (as compared to on-road study) could be the short duration of the experiment, a number of users, and use of different sensors for data collection. Nevertheless, these results show that multi-task learning with reliable quality skin conductance and heart rate signals can be used to detect physiological stress during driving as it generalizes well across various drivers and different environments (real-world and a simulator).

\section{Conclusion}
In this work, the multi-task neural network is used to detect physiological stress during real-world and simulator driving tasks. Generally, a global (subject-independent) model is used for this purpose, which may perform poorly due to large interpersonal variations in physiological parameters (e.g. due to age and diet)~\cite{picard2001toward}. Likewise, most of the studies (see \cite{sharma2012objective} for a review) used sensor data (such as EMG, respiration rate, facial expressions and pupil dilation) that are very hard to acquire in a real-life situation to develop practical applications. Therefore, we used skin conductance and heart rate features in combination with multi-task learning (subjects-as-tasks) to come up with a personalized stress model. In our experiments, we found almost similar results on MIT Driver Stress and Simulator Driving datasets, with a same neural network architecture. Hence, it can be said that if a wearable device provides reliable quality signals, real-time stress detection application can be developed to improve driver's safety and well-being. 

In the future, we will explore transferring representation learned from one dataset to another and to examine the generalization; like it has been usually done for computer vision and natural language processing problems~\cite{sharif2014cnn,chen2012marginalized}. Moreover, we want to apply neural network with temporal convolutions and recurrent layers; on raw physiological signals to automatically learn discriminant features. Most importantly, a future study may involve investigating the performance of these models in real-life situation e.g. by comparing the output of the model against subjective self-reports.  

\medskip

\small{\bibliographystyle{plainnat}
\bibliography{main}}

\begin{thebibliography}{23}
\providecommand{\natexlab}[1]{#1}
\providecommand{\url}[1]{\texttt{#1}}
\expandafter\ifx\csname urlstyle\endcsname\relax
  \providecommand{\doi}[1]{doi: #1}\else
  \providecommand{\doi}{doi: \begingroup \urlstyle{rm}\Url}\fi

\bibitem[Birkett(2011)]{birkett2011trier}
Melissa~A Birkett.
\newblock The trier social stress test protocol for inducing psychological
  stress.
\newblock \emph{Journal of visualized experiments: JoVE}, \penalty0 (56), 2011.

\bibitem[Chen et~al.(2012)Chen, Xu, Weinberger, and Sha]{chen2012marginalized}
Minmin Chen, Zhixiang Xu, Kilian Weinberger, and Fei Sha.
\newblock Marginalized denoising autoencoders for domain adaptation.
\newblock \emph{arXiv preprint arXiv:1206.4683}, 2012.

\bibitem[Dawson et~al.(2007)Dawson, Schell, and
  Filion]{dawson2007electrodermal}
Michael~E. Dawson, Anne~M. Schell, and Diane~L. Filion.
\newblock The electrodermal system.
\newblock \emph{Handbook of psychophysiology}, 2:\penalty0 200--223, 2007.

\bibitem[Ferreira et~al.(2008)Ferreira, Sanches, H{\"o}{\"o}k, and
  Jaensson]{ferreira2008license}
Pedro Ferreira, Pedro Sanches, Kristina H{\"o}{\"o}k, and Tove Jaensson.
\newblock License to chill!: how to empower users to cope with stress.
\newblock In \emph{Proceedings of the 5th Nordic conference on Human-computer
  interaction: building bridges}, pages 123--132. ACM, 2008.

\bibitem[Healey and Picard(2002)]{healey2002driver}
Jennifer Healey and Rosalind~W. Picard.
\newblock Driver stress data.
\newblock \emph{Retrieved June 26th from MIT Affective Computing Group:
  http://affect. media. mit. edu}, 124, 2002.

\bibitem[Healey and Picard(2005)]{healey2005detecting}
Jennifer~A. Healey and Rosalind~W. Picard.
\newblock Detecting stress during real-world driving tasks using physiological
  sensors.
\newblock \emph{IEEE Transactions on intelligent transportation systems},
  6\penalty0 (2):\penalty0 156--166, 2005.

\bibitem[Hennessy and Wiesenthal(1999)]{hennessy1999traffic}
Dwight~A. Hennessy and David~L. Wiesenthal.
\newblock Traffic congestion, driver stress, and driver aggression.
\newblock \emph{Aggressive behavior}, 25\penalty0 (6):\penalty0 409--423, 1999.

\bibitem[Holmes et~al.(2006)Holmes, Krantz, Rogers, Gottdiener, and
  Contrada]{holmes2006mental}
Sari~D. Holmes, David~S Krantz, Heather Rogers, John Gottdiener, and Richard~J.
  Contrada.
\newblock Mental stress and coronary artery disease: a multidisciplinary guide.
\newblock \emph{Progress in cardiovascular diseases}, 49\penalty0 (2):\penalty0
  106--122, 2006.

\bibitem[Jaques et~al.(2016)Jaques, Taylor, Nosakhare, Sano, and
  Picard]{jaquesmulti}
Natasha Jaques, Sara Taylor, Ehimwenma Nosakhare, Akane Sano, and Rosalind
  Picard.
\newblock Multi-task learning for predicting health, stress, and happiness.
\newblock \emph{NIPS Workshop on Machine Learning for Healthcare}, 2016.

\bibitem[Kingma and Ba(2014)]{kingma2014adam}
Diederik Kingma and Jimmy Ba.
\newblock Adam: A method for stochastic optimization.
\newblock \emph{arXiv preprint arXiv:1412.6980}, 2014.

\bibitem[Labb{\'e} et~al.(2007)Labb{\'e}, Schmidt, Babin, and
  Pharr]{labbe2007coping}
Elise Labb{\'e}, Nicholas Schmidt, Jonathan Babin, and Martha Pharr.
\newblock Coping with stress: the effectiveness of different types of music.
\newblock \emph{Applied psychophysiology and biofeedback}, 32\penalty0
  (3-4):\penalty0 163--168, 2007.

\bibitem[Ollander(2015)]{ollander2015wearable}
Simon Ollander.
\newblock Wearable sensor data fusion for human stress estimation, 2015.
\newblock Master Thesis, Technical University of Linköping University.

\bibitem[Picard et~al.(2001)Picard, Vyzas, and Healey]{picard2001toward}
Rosalind~W. Picard, Elias Vyzas, and Jennifer Healey.
\newblock Toward machine emotional intelligence: Analysis of affective
  physiological state.
\newblock \emph{IEEE transactions on pattern analysis and machine
  intelligence}, 23\penalty0 (10):\penalty0 1175--1191, 2001.

\bibitem[Pickering(2001)]{pickering2001mental}
Thomas~G. Pickering.
\newblock Mental stress as a causal factor in the development of hypertension
  and cardiovascular disease.
\newblock \emph{Current hypertension reports}, 3\penalty0 (3):\penalty0
  249--254, 2001.

\bibitem[Ruder(2017)]{ruder2017overview}
Sebastian Ruder.
\newblock An overview of multi-task learning in deep neural networks.
\newblock \emph{arXiv preprint arXiv:1706.05098}, 2017.

\bibitem[Saeed et~al.(2017)Saeed, Trajanovski, van Keulen, and van
  Erp]{saeed2017ddeep}
Aaqib Saeed, Stojan Trajanovski, Maurice van Keulen, and Jan van Erp.
\newblock Deep physiological arousal detection in a driving simulator using
  wearable sensors.
\newblock \emph{IEEE International Conference on Data Mining - workshop: Data
  Mining in Biomedical Informatics and Healthcare (DMBIH)}, 2017.

\bibitem[Schneiderman et~al.(2005)Schneiderman, Ironson, and
  Siegel]{schneiderman2005stress}
Neil Schneiderman, Gail Ironson, and Scott~D. Siegel.
\newblock Stress and health: psychological, behavioral, and biological
  determinants.
\newblock \emph{Annu. Rev. Clin. Psychol.}, 1:\penalty0 607--628, 2005.

\bibitem[Sem-Jacobsen(1961)]{sem1961electroencephalographic}
Carl~W. Sem-Jacobsen.
\newblock Electroencephalographic study of pilot stresses in flight.
\newblock Technical report, Gaustad Hospital Oslo (Norway) Eeg Research Lab,
  1961.

\bibitem[Sexton et~al.(2000)Sexton, Thomas, and Helmreich]{sexton2000error}
J.~Bryan Sexton, Eric~J. Thomas, and Robert~L. Helmreich.
\newblock Error, stress, and teamwork in medicine and aviation: cross sectional
  surveys.
\newblock \emph{British Medical Journal}, 320\penalty0 (7237):\penalty0
  745--749, 2000.

\bibitem[Sharif~Razavian et~al.(2014)Sharif~Razavian, Azizpour, Sullivan, and
  Carlsson]{sharif2014cnn}
Ali Sharif~Razavian, Hossein Azizpour, Josephine Sullivan, and Stefan Carlsson.
\newblock Cnn features off-the-shelf: an astounding baseline for recognition.
\newblock In \emph{Proceedings of the IEEE conference on computer vision and
  pattern recognition workshops}, pages 806--813, 2014.

\bibitem[Sharma and Gedeon(2012)]{sharma2012objective}
Nandita Sharma and Tom Gedeon.
\newblock Objective measures, sensors and computational techniques for stress
  recognition and classification: A survey.
\newblock \emph{Computer methods and programs in biomedicine}, 108\penalty0
  (3):\penalty0 1287--1301, 2012.

\bibitem[Shi et~al.(2007)Shi, Ruiz, Taib, Choi, and Chen]{shi2007galvanic}
Yu~Shi, Natalie Ruiz, Ronnie Taib, Eric Choi, and Fang Chen.
\newblock Galvanic skin response (gsr) as an index of cognitive load.
\newblock In \emph{CHI'07 extended abstracts on Human factors in computing
  systems}, pages 2651--2656. ACM, 2007.

\bibitem[Zhai and Barreto(2006)]{zhai2006stress}
Jing Zhai and Armando Barreto.
\newblock Stress detection in computer users based on digital signal processing
  of noninvasive physiological variables.
\newblock In \emph{Engineering in Medicine and Biology Society, 2006. EMBS'06.
  28th Annual International Conference of the IEEE}, pages 1355--1358. IEEE,
  2006.

\end{thebibliography}

\end{document}